\documentclass[runningheads]{llncs}
\usepackage{comment}
\usepackage{graphicx}
\usepackage{amsfonts}
\usepackage[utf8]{inputenc}
\usepackage[a4paper, left = 2cm, right = 2cm, top=2.5cm,bottom=2.5cm]{geometry}
\usepackage[dvipsnames]{xcolor}
\usepackage{cite}
\usepackage{epstopdf}
\usepackage{subfigure}
\usepackage{mathrsfs}
\usepackage{amsmath}
\usepackage{booktabs}
\usepackage{multirow}
\usepackage{float}
\usepackage{hhline}
\usepackage{algorithm}
\usepackage{algorithmicx}
\usepackage{algpseudocode}
\usepackage{amssymb}
\usepackage{float}
\usepackage{overpic}
\usepackage{pifont}
\usepackage{caption}
\usepackage{marvosym}
\usepackage{ifsym}

\begin{document}
%
\title{DPIT: Dual-Pipeline Integrated Transformer \\for Human Pose Estimation}

\titlerunning{DPIT}

\author{Shuaitao Zhao\inst{1} \and
Kun Liu\inst{2} \and
Yuhang Huang\inst{1} \and
Qian Bao\inst{3} \and
Dan Zeng\inst{1}\textsuperscript{(\Letter)} \and
Wu Liu\inst{3}\textsuperscript{(\Letter)}}

\authorrunning{S. Zhao et al.}
%
\institute{Joint International Research Laboratory of Specialty Fiber Optics and Advanced Communication, Shanghai Institute of Advanced Communication and Data Science, Shanghai University, Shanghai 200444, China \and
JD Logistics, Beijing 100176, China \and
JD Explore Academy, Beijing 100176, China \\
\email{\{zhaoshuaitao,huangyuhang,dzeng\}@shu.edu.cn}\\
\email{\{liukun167,baoqian,liuwu1\}@jd.com}}
\maketitle              

\begin{abstract}
Human pose estimation aims to figure out the keypoints of all people in different scenes. Current approaches still face some challenges despite promising results. Existing top-down methods deal with a single person individually, without the interaction between different people and the scene they are situated in. Consequently, the performance of human detection degrades when serious occlusion happens. On the other hand, existing bottom-up methods consider all people at the same time and capture the global knowledge of the entire image. However, they are less accurate than the top-down methods due to the scale variation. To address these problems, we propose a novel Dual-Pipeline Integrated Transformer (\textbf{DPIT}) by integrating top-down and bottom-up pipelines to explore the visual clues of different receptive fields and achieve their complementarity. Specifically, DPIT consists of two branches, the bottom-up branch deals with the whole image to capture the global visual information, while the top-down branch extracts the feature representation of local vision from the single-human bounding box. Then, the extracted feature representations from bottom-up and top-down branches are fed into the transformer encoder to fuse the global and local knowledge interactively. Moreover, we define the keypoint queries to explore both full-scene and single-human posture visual clues to realize the mutual complementarity of the two pipelines. To the best of our knowledge, this is one of the first works to integrate the bottom-up and top-down pipelines with transformers for human pose estimation. Extensive experiments on COCO and MPII datasets demonstrate that our DPIT achieves comparable performance to the state-of-the-art methods.

\end{abstract}
\section{Introduction}
Human Pose Estimation (HPE) has been widely investigated as a fundamental task in computer vision, which aims to localize keypoints of the human, including eyes, nose, shoulders, wrists, \textit{etc.}, from a single RGB image. Accurate human pose estimation can provide geometric and motion information about the human, which can be widely applied in action recognition \cite{liu2018t,liu2020real}, human-computer interaction, motion analysis, augmented reality (AR), \textit{etc.}

Early human pose estimation methods do not depend on deep learning and mainly focus on the keypoints localization of a single person, which can be roughly divided into two categories. The first category treats the pose estimation task as a classification or regression problem through a global feature \cite{randomized2008,sparse2008}. However, this kind of method does not exhibit high precision and is only suitable for clean scenes. The other category is the methods adopt graphic model to extract the feature representation for a single keypoint \cite{strong2013,latent2011}. The location of a single part can be obtained using DPM (Deformable Part-based Model) \cite{DPM2010}, and the pair-wise relationships are required to optimize the association between keypoints at the same time.

Recently, with the rapid development of deep learning, Convolutional Neural Networks (CNNs) have shown strong dominance in human pose estimation. We can roughly classify these superior networks into top-down and bottom-up methods. The top-down methods \cite{CPM2016,Hourglass2016,CPN2018,simple_baseline2018,HRNet2019,MSPN2019,graph2020,learning2020} first obtain a set of the bounding box of people from the input image through an off-the-shelf human detector , then apply a single-person pose estimator to each person. This type of method mainly focuses on the investigation and improvement of the latter pose estimation network. Different from the top-down pipeline, the bottom-up methods \cite{deepcut2016,0penpose2017, RMPE2017, Higherhrnet2020,Simple2021} directly predict all the joints in an image and then group them using a certain assignment strategy to achieve multi-person pose estimation.

The top-down methods rely on the result of human detection and achieve promising performance for single-person pose estimation. However, because they deal with each person individually, there is no awareness of the interaction with the other persons and the environment, which is more prevalent in real-life scenarios. When there is serious occlusion among different people, the performance of human detection becomes unreliable.
\begin{figure*}
  \centerline{\includegraphics[width=0.8\textwidth]{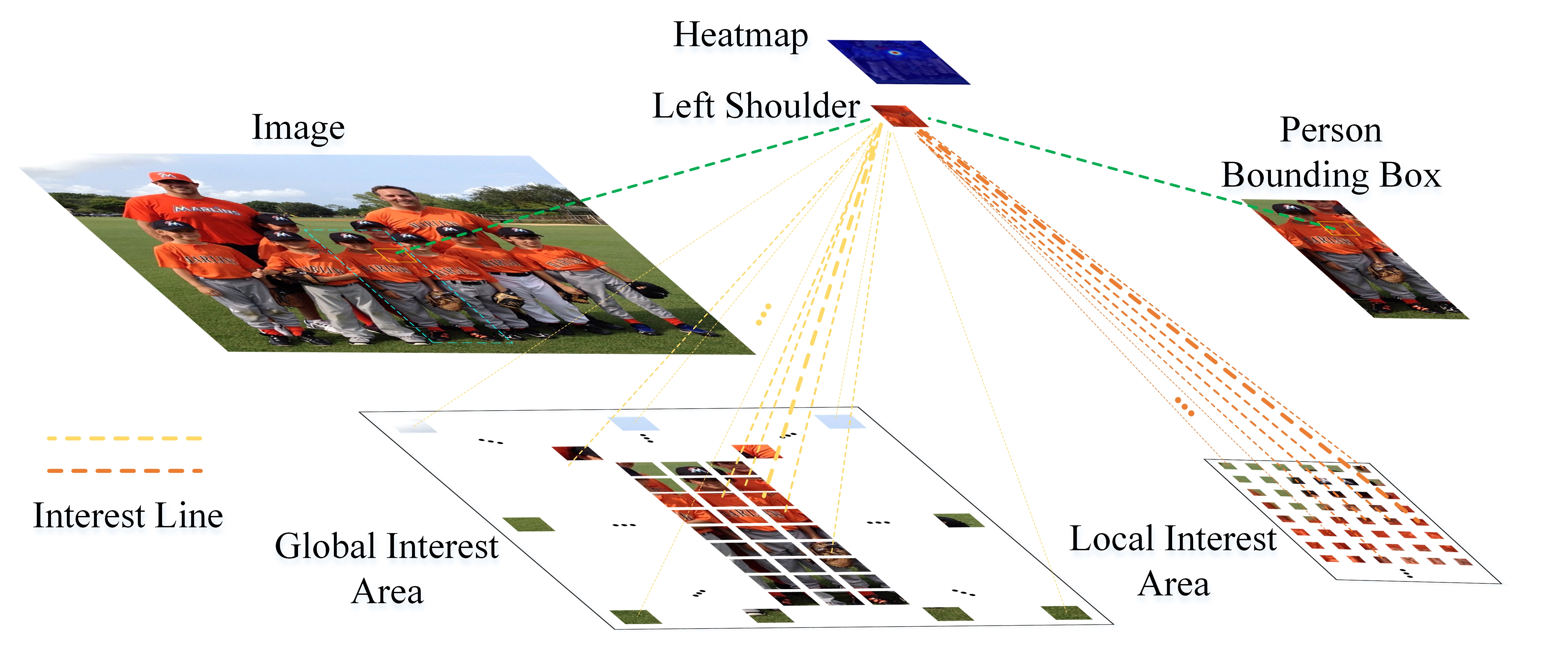}}
  \caption{Illustration of capturing information from different perspectives when predicting the left shoulder. The dotted lines point to the interest areas of the keypoint, and the thicker line indicates the more interest to the area. After the integration of different visual clues, the keypoint location is predicted by the heatmap.}
  \label{figs_1}
\end{figure*}
Furthermore, when the target persons are very close to each other, the pose estimator may be misled by nearby persons, e.g., the predicted keypoints may come from adjacent persons. As a result, the top-down pipeline exhibits an inherent limitation in how to explore the interaction clues among different persons. Differently, the bottom-up methods do not rely on any detection process and
take all people of the image into account simultaneously. They first detect the keypoints of all people, then align them into each person by a certain grouping strategy. This pipeline leverages full-scene information to realize locating keypoints, which can observe the interactions of different people from a global perspective. However, it suffers from scale variation, i.e., different people in the image are at different scales and very unevenly distributed, which is unfriendly to the training of the network, leading to relatively poor performance. In summary, both kinds of approaches 
show different advantages. Consequently, integrating the advantages of the two pipelines is potential for human pose estimation.

To achieve the complementarity of two pipelines, we propose an effective network called DPIT to further promote the visual exploration of the image for human pose estimation. The proposed network integrates the advantages of the top-down and bottom-up pipelines to learn long-range visual clues with different receptive fields, which capture the full-scene and posture information of a single person. Firstly, we design two branches to extract global scene features and local features of a single person, respectively. Secondly, to fuse these features, we employ the transformer to capture different visual cues. The final output of our DPIT is predicted in the heatmap fashion.

Since the two pipelines of features contained different information are both essential for keypoints localization, our core idea is to incorporate the two pipelines into one network and perform effective information interaction. An example of how to predict the location of the left shoulder is shown in Fig.~\ref{figs_1}. Information from different perspectives can assist the network in better understanding the image scenes, and extend the interest region, which is of positive effects for human pose estimation in complex scenes.
More specifically, we first employ a two-branch structure, in which the bottom-up branch captures full-scene information of the entire image, the top-down branch extracts the single-person feature of the detected human bounding box. Motivated by the great success of the recent work on Vision Transformers, the encoder of the transformer is employed as the clue interaction structure to fuse the two-branch features. In detail, we split the features into patches and take linear patch embeddings to form input visual tokens. Meanwhile, we define a set of randomly initialized embeddings as the keypoint queries, which can capture the single-person pose, full-scene visual clues, and their distributional relations from the different tokens. Finally, we only apply the heatmap generator to the keypoint embeddings which have aggregated local and global information, and reshape them into heatmaps. To the best of our knowledge, this is one of the first works to integrate the two pipelines with transformers.

In summary, the contributions of this paper are mainly summarized as below:
\begin{itemize}
    \item We propose a novel architecture named DPIT for human pose estimation, which is one of the first works to integrate the bottom-up and top-down pipelines in an end-to-end training manner.
    \item We design a Transformer-based module to capture both full-scene visual clues and posture information of a single person simultaneously, which can allocate different attention levels to different visual areas.
    \item We demonstrate an improvement over baselines on the widely used COCO and MPII datasets, and surpass the state-of-the-art methods.
\end{itemize}

The rest of this paper is organized as follows. In Sec.~\ref{sec:RelatedWork}, we review the related works. In Sec.~\ref{sec:Method}, we introduce the proposed DPIT for human pose estimation. Extensive experiments are conducted in Sec.~\ref{sec:Experiments} to compare the proposed DPIT with state-of-the-art methods on two benchmark datasets. The conclusion is given in Sec.~\ref{sec:Conclusion}.

\section{Related Work} \label{sec:RelatedWork}
Our proposed method is related to the previous research on top-down human pose estimation, bottom-up human pose estimation, and applications of transformer in vision tasks.

\subsection{Top-down Human Pose Estimation}
The Top-down pipeline consists of two main components: the human detector and the pose estimation network. Most of the work \cite{CPN2018,MSPN2019,simple_baseline2018,HRNet2019,RSN2020,graph2020} focused on the design and improvement of the latter pose estimation network. CPN \cite{CPN2018} implemented the coarse-to-fine process through a two-stage network, where the GlobalNet learns a well-defined feature representation based on a feature pyramid network to provide sufficient semantic information to locate simple keypoints. Further, the RefineNet is employed to handle the ‘‘difficult'' keypoints by fusing the multi-level features of the GlobalNet. MSPN \cite{MSPN2019} performed stacking of multiple stages based on CPN's globalNet to achieve better information communication. Xiao \textit{et al.} \cite{simple_baseline2018} employed ResNet as the backbone and added some de-convolution layers behind it, which built a simple but effective structure to produce a high-resolution representation of the keypoint heatmap. HRNet \cite{HRNet2019} started to give attention to the importance of spatial resolution. A novel high-resolution network is proposed to learn the reliable high-resolution features by connecting multi-resolution sub-networks in parallel, as well as performing repetitive multi-scale fusion. Cai \textit{et al.} \cite{RSN2020} proposed a multi-stage network where the Residual Step Network (RSN) explores delicate local features through an effective inter-level feature fusion strategy. In addition, a new attention mechanism (PRM) was also proposed to learn different contributions for local and global features, achieving more accurate keypoint localization. Wang \textit{et al.} \cite{graph2020} proposed a graph-based, model-independent two-stage network, Graph-PCNN. This framework added a localization sub-network and a graph structure pose optimization module to the original heatmap-based regression method. The heatmap regression network was employed as the first stage to provide rough localization of each keypoint. The localization sub-network was designed as the second stage to extract visual features from the candidate keypoints. Although these top-down methods can achieve satisfactory performance for the single-person bounding box, they are unreliable in obscured scenes. 

\subsection{Bottom-up Human Pose Estimation}
The bottom-up pipeline consists of two main stages, including the joints detection and grouping of all human keypoints in the image \cite{0penpose2017,associative2017,RMPE2017,Higherhrnet2020,Simple2021,DEKR2021}. OpenPose \cite{0penpose2017} predicted the heatmap of keypoints to locate the position of each keypoint in the image. The Part Affinity Field (PAF) was proposed to achieve the connection of keypoints, which speeds up the bottom-up human pose estimation to a great extent. Associative Embedding \cite{associative2017} not only predicted the heatmaps but also output an embedding for each keypoint, aiming to make the embeddings of the same person as similar as possible. RMPE \cite{RMPE2017} proposed a two-step framework, which mainly solved the positioning error and the redundancy of the bounding box. HigherHRNet \cite{Higherhrnet2020} provided a simple extension to HRNet \cite{HRNet2019} by deconvoluting the high-resolution heatmap to obtain the higher resolution representation. SIMPLE \cite{Simple2021} employed knowledge distillation by treating the top-down network as a teacher network to train the bottom-up network. Both human detection and keypoint estimation were considered as unified point learning issues that complement each other in a single framework. DEKR \cite{DEKR2021} proposed a simple but effective method that employs adaptive convolution through a pixel-wise spatial transformer to activate pixels in the keypoint regions. Separate regression of different keypoints was also performed using a multi-branch structure. The separated representations can notice the keypoint regions separately so that the keypoint regression is more spatially accurate. However, the variation of the person scale in the image significantly affects the performance of these bottom-up methods.

\subsection{Transformers in Vision}
The amazing achievements of the transformer in natural language have attracted the vision community to explore its application to computer vision tasks. Recently, the transformer has been widely applied in different vision tasks including image classification \cite{VIT2020}, object detection \cite{DETR2020,deformable2020}, segmentation \cite{segformer2021,segmenter2021}, and generation \cite{sceneformer2021}, \textit{etc}.

\begin{figure*}
  \centering
  \includegraphics[width=\textwidth]{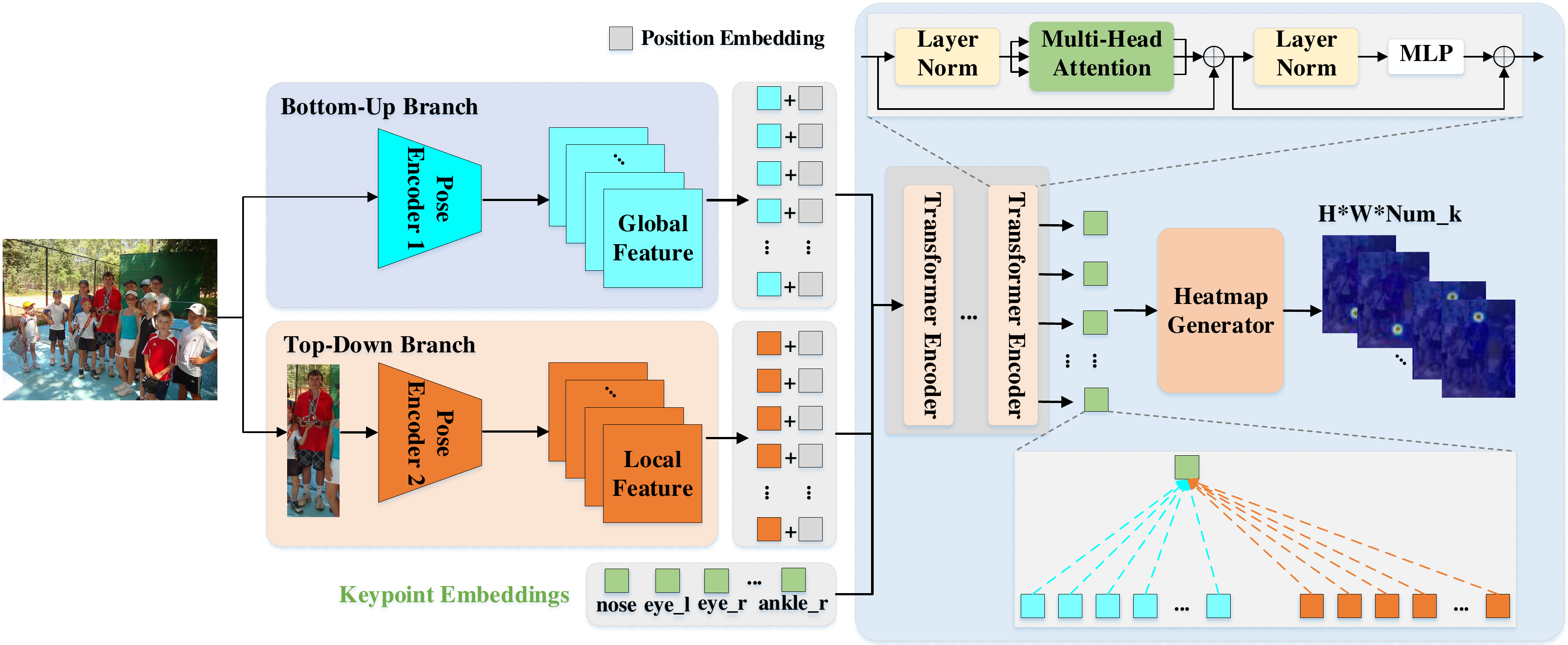} 
  \caption{The overall training architecture of our network. The image is input to the bottom-up branch to get the full-scene feature. The single-person bounding box output from the human detector is fed into the top-down branch to extract the single-human pose feature. Then the defined random embeddings are treated as the keypoint queries, which is sent into the transformer encoder together with the visual tokens in sequential fashion. The outputs of our network are the heatmaps of keypoints with the shape of $H \times W \times Num\_k$, where $Num\_k$ is the number of keypoints. All components of the network are trained together in an end-to-end manner.}
  \label{overall_architecture}
\end{figure*}

ViT \cite{VIT2020} completely abandoned CNN and applied transformer to image classification, which splits the image into fix-sized patches, each of which is expanded into sequential form and fed to the encoder of transformer by linear projection. DeiT \cite{DEiT2021} incorporated distillation into the training of ViT. It introduced a teacher-student training strategy, in which the convolution network was employed as the teacher network. 
DETR \cite{DETR2020} applied transformer to the object detection. The image is processed by CNN for feature extraction, then fed into the transformer in the manner of feature sequences. The transformer architecture directly outputs an unordered set. Each element of the set contains object categories and coordinates. SegFormer \cite{segformer2021} proposed a simple and efficient structure for semantic segmentation, consisting of a positional-encoding-free, hierarchical transformer encoder and an MLP decoder, which achieves high efficiency and accuracy.

In human pose estimation, the transformer has also received extensive attention and application \cite{POET2021,Tfpose2021,Transpose2021,Tokenpose2021,Swinpose2022}. POET \cite{POET2021} proposed an encoder-decoder structure combining CNN and transformer, which can directly regress the pose of all individuals using a bipartite matching scheme. Based on the regression manner, TFPose \cite{Tfpose2021} implemented direct human pose estimation, overcoming the feature-mismatch problem of previous regression-based methods. TransPose \cite{Transpose2021} introduced a transformer-based structure to predict the location of human keypoints based on heatmaps, which can effectively capture the spatial relationships of images. Following ViT \cite{VIT2020}, TokenPose \cite{Tokenpose2021} divided the image into several patches to form the visual tokens, which incorporated the visual cue and constraint cue into a unified network.
Swin-Pose \cite{Swinpose2022} proposed a transformer-based structure to capture the long-range dependencies between pixels, using the pre-trained Swin Transformer \cite{swin-transformer2021} as the backbone to extract image features. In addition, the feature pyramid architecture was adopted to fuse features from different stages for feature enhancement. Different from these transformer-based methods, our method employ the transformer to integrate the top-down and bottom-up pipelines. In this way, our network can capture the global clues and local clues simultaneously.

\section{The Proposed Method} \label{sec:Method}
The presented framework is shown in Fig.~\ref{overall_architecture}, where all modules are trained in an end-to-end manner. Given an image as input, a two-branch architecture is employed for feature extraction, where the bottom-up branch and top-down branch extract full-scene information and posture of a single person, respectively. Then, a transformer-based integration network is designed to jointly establish keypoint-person and keypoint-scene interactions.

\subsection{Two-branch Architecture}
As shown in Fig.~\ref{overall_architecture}, to integrate the top-down and bottom-up pipelines into a unified network, we employ a parallel two-branch CNN for feature extraction in the first stage of our network. The two CNN backbones are pre-trained on the ImageNet classification task \cite{ImageNet2015}. Specifically, an image $x_{1} \in {{\mathbb{R}}^{H_{1} \times W_{1} \times 3}}$ is first detected by an existing human detector, from which we obtain the human bounding box $x_{2} \in {{\mathbb{R}}^{H_{2} \times W_{2} \times 3}}$. 
The bottom-up encoder $\textbf{\textit{E}}_{}^{BU}$ extracts the full-scene feature representation ${\textbf{\textit{F}}_{}^{BU} \in {{\mathbb{R}}^{{H_{1}^{'}} \times {W_{1}^{'}} \times {C^{}}}}}$ from the input image $x_{1}$. Taking $x_{2}$ as input, the top-down encoder $\textbf{\textit{E}}_{}^{TD}$ outputs the posture representation ${\textbf{\textit{F}}_{}^{TD} \in {{\mathbb{R}}^{{H_{2}^{'}} \times {W_{2}^{'}} \times {C^{}}}}}$ of single-person. The process of feature extraction of the two branches can be represented as:
\begin{equation}
\label{equ1}
    \textbf{\textit{F}}_{}^{BU} = \textbf{\textit{E}}_{}^{BU}({x_1}), \textbf{\textit{F}}_{}^{TD} = \textbf{\textit{E}}_{}^{TD}({x_2}),
\end{equation}
where $\textbf{\textit{E}}_{}^{BU}$ and $\textbf{\textit{E}}_{}^{TD}$ denote bottom-up and top-down CNN encoders, respectively.

The featuremap extracted by the bottom-up encoder comes from the whole image, where we can observe the full-scene information, including the pose of all people, the interaction of different people, and the scene information of the image. The top-down branch focuses on the spatial posture information of a single person with uniform resolution. In real scenes, the environment is usually undefinable, there are various interactions among different people. As a result, capturing visual clues from different view fields can assist in locating the keypoints more effectively.

\subsection{Transformer-based Integration}
Inspired by the wide applications of transformer in vision tasks, we employ the encoder of the transformer to capture and integrate the visual clues of different view fields. 
Together with the patch embeddings from the bottom-up and top-down branches, the keypoint embeddings are defined as input queries. In this way, our network can capture the knowledge of global fields and local fields that comes from the whole image and single-human bounding box, respectively.

\subsubsection{Input Queries}
Following the process of ViT \cite{VIT2020}, The featuremap $\textbf{\textit{F}}_{}^{BU}$ and $\textbf{\textit{F}}_{}^{TD}$ are split into two patch sets: $p_{1} = [\textit{F}_{1}^{BU}, ..., \textit{F}_{N_{1}}^{BU}] \in {{\mathbb{R}}^{{N_{1}} \times {P_{1}^{h} \times P_{1}^{w}} \times {C^{}}}}, p_{2} = [\textit{F}_{1}^{TD}, ..., \textit{F}_{N_{2}}^{TD}] \in {{\mathbb{R}}^{{N_{2}} \times {P_{2}^{h} \times P_{2}^{w}} \times {C^{}}}}$, where ($P_{*}^{h}, P_{*}^{w}$) ($* = 1, 2$) is the patch size, $N_{*} = (H_{*}^{'} \times W_{*}^{'}) / (P_{*}^{h} \times P_{*}^{w})$ is the number of patches, $C^{}$ is the number of channels. The standard Transformer \cite{transformer2017} receives a 1D sequence of the token embeddings as input. Every patch is flattened into a 1D vector with the size of $P_{*}^{h} \cdot P_{*}^{w} \cdot C^{}$. Then, the linear projection is performed to each 1D vector, we can get the visual queries of two branches: $\mathbf{q_{1}}$ = $[E_{1}, ..., E_{N_{1}}]$, $\mathbf{q_{2}} = [E_{1}, ..., E_{N_{2}}], $where $E \in {{\mathbb{R}}^{{(P_{*}^{h} \cdot P_{*}^{w} \cdot C^{}) \times D}}}$ and $D$ is the embedding dimension. It is worth mentioning that human pose estimation is a location sensitive vision task, so the 2D position embeddings are added to the sequence of patches to capture positional information.

Moreover, we introduce a set of learnable keypoint embeddings: $\textbf{kpt} = [k_{1}, ..., k_{K}] \in {\mathbb{R}}^{{K \times D}}$, where $K$ is the number of keypoints. Each keypoint embedding is initialized randomly and assigned to a single keypoint (eyes, wrists, ...), which is employed to generate the final heatmap. Benefit from the self-attention mechanism of transformer, each keypoint embedding can capture the corresponding interest visual regions from both global and local features. In addition, it can give attention to the clues of other keypoints, which simplifies the difficulty of locating keypoints, especially in complex scenarios. The keypoint embeddings \textbf{kpt}, bottom-up visual query $\mathbf{q_{1}}$, and top-down visual query $\mathbf{q_{2}}$ are put together as the input queries, which are processed jointly by the transformer encoder, as depicted in Fig.~\ref{overall_architecture}. 

\subsubsection{Transformer Encoder}
The transformer encoder consists of multi-encoder layers, which mainly depend on the self-attention mechanism. Each encoder layer contains a multi-head self-attention ($MSA$) block followed by a feed-forward network ($FFN$). Layer Normalization ($LN$) is also employed before every module and residual connections after every module. Specifically, for the input sequence X, linear projections are performed to obtain Query ($Q$), Key ($K$), and Value ($V$):
\begin{equation}
    Q = X * W_{Q}, K = X * W_{K}, V = X * W_{V},
\end{equation}
where $W_{Q}$, $W_{K}$, $W_{V}$ are the corresponding weight matrices. The $MSA$ process can formulated:
\begin{equation}
    MSA(Q,K,V) = softmax(\frac{{Q \times {K^T}}}{{\sqrt {{{\rm{d}}_k}} }}) \cdot V,
\end{equation}
where $d_{k}$ is the dimension of keys. Each query is calculated with all the keys, each ($Q$, $K$) pair is divided by ${{{\rm{d}}_k}}$. Then, the SoftMax function is employed to obtain the attention scores, each score determines the attention level to the token for current query. 

\subsubsection{Heatmap Generator}
The transformer encoder outputs a D-dimensional sequence. After the transformer's self-attention mechanism, the corresponding keypoint embeddings have captured the visual clues of the full-scene and single-person. We only take the keypoint queries for prediction, which are linearly mapped into $H\cdot W$ dimensions. Then the mapped 1D representations are reshaped into 2D heatmaps with the shape of $H \times W$. Finally, on output heatmaps, we find the maximum response position by channel to locate the corresponding human keypoint. In addition, between the output heatmap and the ground-truth heatmap, we adopt MSE as the loss function to train the network.

\section{Experiments} \label{sec:Experiments}
\subsection{Datasets}
In this paper, we conducted extensive experiments on two human pose datasets, COCO \cite{COCO2014microsoft} and MPII \cite{MPII20142d}, to train and validate our network. For a fair comparison, we follow the same dataset split ratio as the comparison methods~\cite{HRNet2019}. The datasets are introduced below.

\subsubsection{COCO Dataset}
COCO is a large-scale dataset in human pose estimation task, containing over 200K images and 250K person instances, annotated with 17 keypoints. The dataset is divided into a train set (118k images), a validation set (5K images), and a test-dev set (20K images). We take the train set to train our network. The validation set and test-dev set are employed to measure the performance of our network.

\subsubsection{MPII Dataset} 
MPII is a well-known benchmark for the evaluation of human pose estimation, which contains around 25K images and over 40K people with annotated 16 joints.

\subsection{Metrics}
For the COCO dataset, the Object Keypoint Similarity ($OKS$) is calculated for the reported metrics, which measures the similarity between the ground truth and predicted keypoints. The $OKS$ is defined as follows:
\begin{equation}
    OKS = \frac{{\sum\nolimits_i {\exp (\frac{{ - d_i^2}}{{2{s^2}k_i^2}})\delta ({v_i} > 0)} }}{{\sum\nolimits_i {\delta ({v_i} > 0)} }},
\end{equation}\
where $d_i$ is the Euclidean distances between each corresponding ground truth and detected keypoint. $v_i$ is the visibility flag of the ground truth. $s$ denotes the square root of the person's proportion to the image area, $\delta_{i}$ is the normalized parameter of the ith keypoint. Based on $OKS$, the standard average precision and average recall are reported, including $AP$ (the mean value of $AP$ at $OKS$ = 0.5, 0.55, ..., 0.9, 0.95), $AP^{50}$ ($AP$ at $OKS$ = 0.5), $AP^{75}$ ($AP$ at $OKS$ = 0.75), $AP^{M}$ ($AP$ of medium-scale objects), $AP^{L}$ ($AP$ of large-scale objects), and $AR$ (mean value of $AR$ at $OKS$ = 0.5, 0.55, ..., 0.9, 0.95).

For MPII, the head-normalized Percentage of Correct Keypoints (PCKh@0.5) \cite{MPII20142d} is employed as the metric, which calculates the percentage of the normalized distance between the ground truths and detected keypoints that are lower than the setting threshold (0.5).

\subsection{Implementation Details}
\label{sec:Implementation}
Regarding the training scheme, all modules of our network are trained with adaptive moment estimation optimizer (ADAM) \cite{2014Adam}, whose parameters are: $\alpha=0.001$, $\beta_{1}=0.9$, $\beta_{2}=0.999$. The initial learning rate is $10^{-3}$ for our network, which is trained for a total of 240 epochs. The learning rate is reduced to $10\%$ of the previous number at the 190th and 220th epochs, respectively. For the backbones of the bottom-up and top-down branches, we adopt the models trained on the ImageNet classification task as our pre-trained models. To improve the varieties of training data, following \cite{HRNet2019}, the data augmentations are conducted, including random rotation ([$-45^{\circ}$, $45^{\circ}$]), random scale ([0.65, 1.35]) and flipping.

\begin{center}
\begin{table}
  \caption{The network configurations. HRNet-W32-s and HRNet-W48-s denote the first three stages of HRNet-W32 and HRNet-W48, respectively.}
  \label{Setting_Table}
  \begin{center}
  \begin{tabular}{c|c|ccc}
  \toprule
  Model     & Backbone  & Depth  & Heads  & Hidden\_Dim \\
  \midrule
  \midrule
  DPIT-B    & HRNet-W32-s  & 12  & 8  & 192     \\
  DPIT-L    & HRNet-W48-s  & 12  & 8  & 192   \\
  \bottomrule
  \end{tabular}
  \end{center}
\end{table}
\end{center}

\begin{center}
\begin{table}
  \caption{Quantitative results on COCO validation set across various state-of-the-art methods with the ground-truth bounding boxes. R and H denote the ResNet and HRNet, respectively. $\#$Params indicates the size of each model, excluding the cost of the human detection network. In each column, the best result is in bold, the second best is underlined.}
  \label{Results_coco}
  \begin{center}
  \begin{tabular}{c|c|c|c|cccccc}
  \toprule
  Method   & Backbone  & Input Size  & $\#$Params  & $AP$ & $AP^{50}$  & $AP^{75}$  & $AP^{M}$  & $AP^{L}$ & $AR$ \\
  \midrule
  \midrule
  \multirow{6}*{Simple Baseline \cite{simple_baseline2018}}    
                            & R-50 & $256 \times 192$  & 34.0M  & 72.4 & 91.5  & 80.4 &  69.7  & 76.5 & 75.6    \\
                            & R-50 & $384 \times 288$  & 34.0M  & 74.1 & 92.6  & 80.5  & 70.5  & 79.6 & 76.9     \\
                            & R-101  & $256 \times 192$  & 53.0M  & 73.4 & 92.6  & 81.4  & 70.7  & 77.7 & 76.5    \\
                            & R-101  & $384 \times 288$  & 53.0M  & 75.5 & 92.5  & 82.6  & 72.4  & 80.8 & 78.4    \\
                            & R-152  & $256 \times 192$  & 68.6M  & 74.3 & 92.6  & 82.5  & 71.6  & 78.7 & 77.4    \\
                            & R-152  & $384 \times 288$  & 68.6M  & 76.6 & 92.6  & 83.6  & 73.7  & 81.3 & 79.3    \\
  \midrule
  \multirow{2}*{HRNet \cite{HRNet2019}}              
                                    & H-W32  & $256 \times 192$  & 28.5M  & 76.5  & \underline{93.5} & 83.7  & 73.9  & 80.8  & 79.3    \\
                                    & H-W48  & $256 \times 192$  & 63.6M  & \underline{77.1} & \textbf{93.6}  & \underline{84.7}  & \underline{74.1}  & \underline{81.9} & \underline{79.9}    \\
  \midrule
  DPIT-B & -  & $256 \times 192$  & 20.8M  & 76.9 & \underline{93.5}  & 83.7  & 73.7  & 81.5 & 79.6    \\
  DPIT-L & -  & $256 \times 192$  & 38.0M  & \textbf{77.8} & \textbf{93.6}  & \textbf{84.8}  & \textbf{74.8}  & \textbf{82.2} & \textbf{80.3}    \\
  \bottomrule
  \end{tabular}
  \end{center}
\end{table}
\end{center}

While training, for the bottom-up branch, we resize the image to a fixed size: $512 \times 512$, then fed it into the encoder of this branch. As introduced before, we split the featuremap output from the encoder into patches with the size of $8 \times 8$. On the other hand, the input of the top-down branch is also rescaled into fixed resolution. There are different settings for COCO and MPII datasets. On COCO, the patch size is set to $4 \times 3$ with input size of $256 \times 192$. On MPII, with a uniform input shape of $256 \times 256$, the feature is split into $4 \times 4$ patches. In addition, our network set up two configuration versions, DPIT-B and DPIT-L. The detailed settings of them are shown in Table~\ref{Setting_Table}.

\subsection{Quantitative Results} \label{sec:eval}
To validate the effectiveness and superiority of our method, we conducted quantitative experiments on COCO and MPII datasets. The quantitative results show that our method achieves better performance on human pose estimation, the specific results are analyzed as follows:

\subsubsection{Results on COCO Dataset}
The quantitative results using the ground-truth bounding box on the COCO validation set are shown in Table~\ref{Results_coco}. For the different methods, we perform quantitative comparisons based on different backbones and input resolutions. Following TokenPose~\cite{Tokenpose2021}, we do not employ the whole HRNet as our backbone, but its first three stages. In this case, the network parameters are only 25\% of the original version, as indicated by HRNet-W32-s and HRNet-W48-s in Table~\ref{Setting_Table}. Compared to SimpleBaseline \cite{simple_baseline2018} and HRNet \cite{HRNet2019}, it can be observed that our method achieves better performance while being more lightweight. Quantitatively, our DPIT-L achieves improvements of 0.7 $AP$ and 0.4 $AR$ compared to the HRNet-W48, which demonstrates the superiority of our method.

In addition, as shown in Table~\ref{test_coco}, we compare our method with the state-of-the-art methods including G-RMI \cite{G-RMI2017}, CPN\cite{CPN2018}, RMPE \cite{RMPE2017}, SimpleBaseline \cite{simple_baseline2018} and HRNet \cite{HRNet2019} on COCO test-dev set. Compared with other methods, DPIT exhibits the best performance on $AP$ and $AR$, achieves comparable results on other metrics.

\begin{center}
\begin{table}
  \caption{Comparison with various state-of-the-art methods with detected bounding boxes from the same human detector on COCO test-dev set.}
  \label{test_coco}
  \begin{center}
  \begin{tabular}{c|c|c|cccccc}
  \toprule
  Method  & Input Size  & $\#$Params  & $AP$ & $AP^{50}$  & $AP^{75}$  & $AP^{M}$  & $AP^{L}$ & $AR$ \\
  \midrule
  \midrule
   G-RMI \cite{G-RMI2017}  & $353 \times 257$  & 42.6M  & 64.9 & 85.5  & 71.3 &  62.3  & 70.0 & 69.7    \\
   CPN \cite{CPN2018}  & $384 \times 288$  & 45.0M  & 72.1 & 91.4  & 80.0  & 68.7  & 77.2 & 78.5    \\
  RMPE \cite{RMPE2017}   & $320 \times 256$  & 28.1M  & 72.3  & 89.2 & 79.1  & 68.0  & \textbf{80.8}  & 78.6    \\
  SimpleBaseline-R152 \cite{simple_baseline2018}   & $384 \times 288$  & 68.6M  & 73.7  & \underline{91.9} & 81.1  & 70.3  & 80.0  & 79.0    \\
  HRNet-W48 \cite{HRNet2019}   & $256 \times 192$  & 63.6M  & \underline{74.2} & \textbf{92.4}  & \textbf{82.4}  & \underline{70.9}  & 79.7 & \underline{79.5}    \\
  \midrule
  DPIT-B   & $256 \times 192$  & 20.8M  & 73.6 & 91.4  & 81.2  & 70.4  & 79.5 & 78.9    \\
  DPIT-L   & $256 \times 192$  & 38.0M  & \textbf{74.6} & \underline{91.9}  & \underline{82.1}  & \textbf{71.3}  & \underline{80.6} & \textbf{79.9}    \\
  \bottomrule
  \end{tabular}
  \end{center}
\end{table}
\end{center}

\subsubsection{Results on MPII Dataset}
The PCKh@0.5 results on the MPII validation set are reported in Table~\ref{Results_mpii} with a uniform input size of $256 \times 256$. DPIT-L/D6 represents the configured DPIT-L with 6 transformer encoder layers. Specifically, compared with SimpleBaseline \cite{simple_baseline2018} and HRNet \cite{HRNet2019}, our DPIT-L/D6 achieves the best performance on the metrics reported by $Elb, Wri, Ank$, and $Mean$. On most other metrics, it also achieves the second-best level. In summary, the quantitative results indicate the comparable performance of our DPIT on the MPII dataset.

\begin{center}
\begin{table}
  \caption{Quantitative Results on MPII validation set. Experiments for all architectures are performed at the uniform input size: $256 \times 256$.}
  \label{Results_mpii}
  \begin{center}
  \begin{tabular}{c|c|c|cccccccc}
  \toprule
  Method     & Backbone & $\#$Params  & Hea & Sho & Elb & Wri & Hip & Kne & Ank & Mean   \\
  \midrule
  \midrule
  \multirow{3}*{SimpleBaseline \cite{simple_baseline2018}}    
                                    & R-50 & 34.0M & 96.4 & 95.3 & 89.0 & 83.2 & 88.4 & 84.0 & 79.6 & 88.5    \\
                                    & R-101  & 53.0M & 96.9 & \underline{95.9} & 89.5 & 84.4 & 88.4 & 84.5 & 80.7 & 89.1    \\
                                    & R-152  & 68.6M & \underline{97.0} & \underline{95.9} & 90.0 & 85.0 & \underline{89.2} & 85.3 & \underline{81.3} & \underline{89.6}    \\
  \midrule
  HRNet \cite{HRNet2019} & H-W32  & 28.5M & 96.9 & \textbf{96.0} & \underline{90.6} & \underline{85.8} & 88.7 & \textbf{86.6} & \textbf{82.6} & \textbf{90.1}    \\
  \midrule
     DPIT-B  & -  & 21.6M  & \textbf{97.1}  & 95.7 & 90.0  & 84.6  & \textbf{89.4}  & 85.9 & 80.7 & \underline{89.6}    \\
     DPIT-L/D6  & -  & 31.8M & 96.7  & \underline{95.9}  & \textbf{90.8} & \textbf{85.9}  & \underline{89.2}  & \underline{86.0}  & \textbf{82.6} & \textbf{90.1}    \\
  \bottomrule
  \end{tabular}
  \end{center}
\end{table}
\end{center}

\begin{figure*}
  \includegraphics[width=\textwidth]{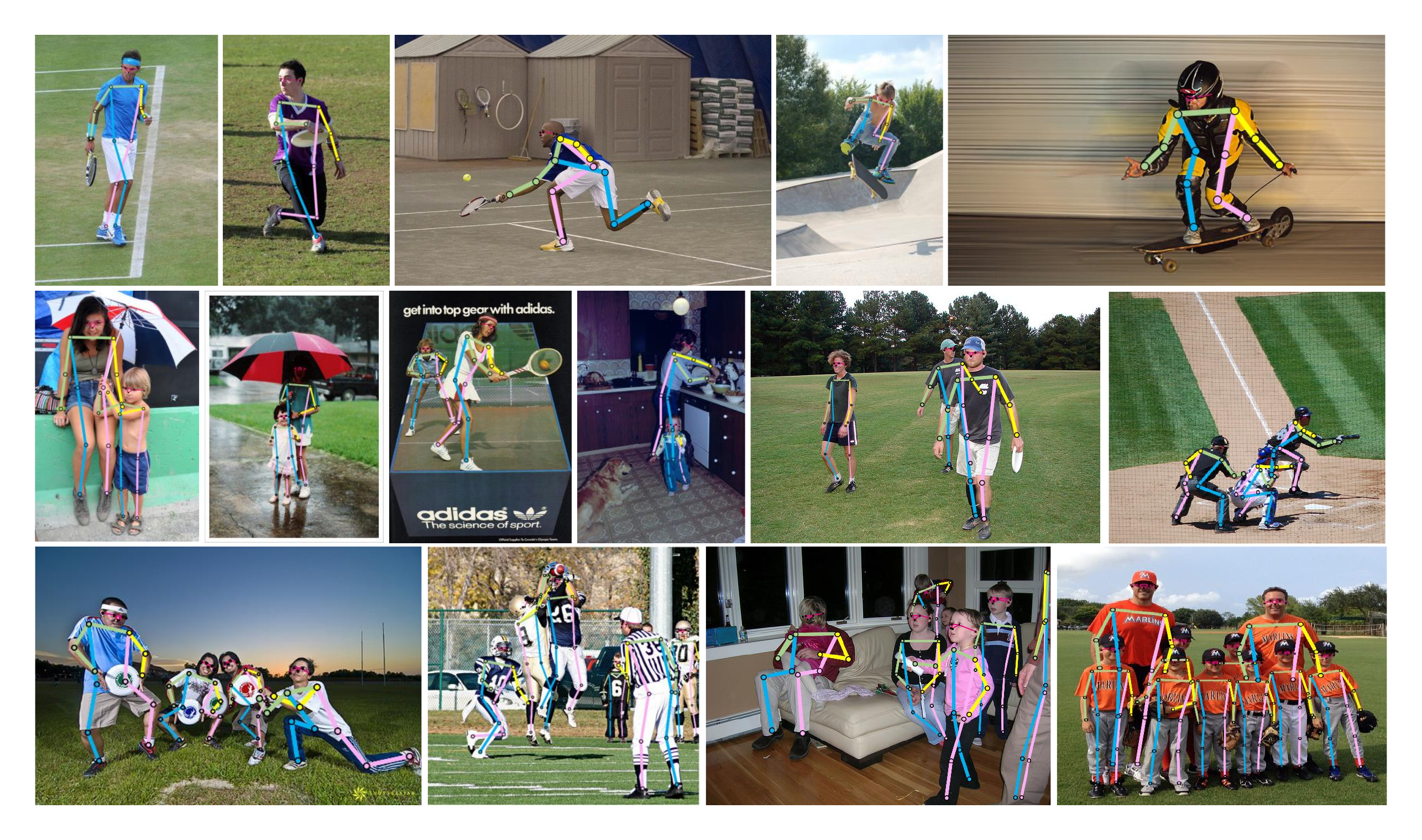}
  \caption{Illustration of human pose estimation results of DPIT in different scenes on COCO validation set. The first row shows the effect of pose estimation in single-person situations with different postures. The scenes in the second row contain interactions of different people, including self-shadowing and inter-shadowing between two persons.  Finally, more complex scenes with multiple people are further visualized in the third row.}
  \label{qual_figs}
\end{figure*}

\begin{figure*}
  \includegraphics[width=\textwidth]{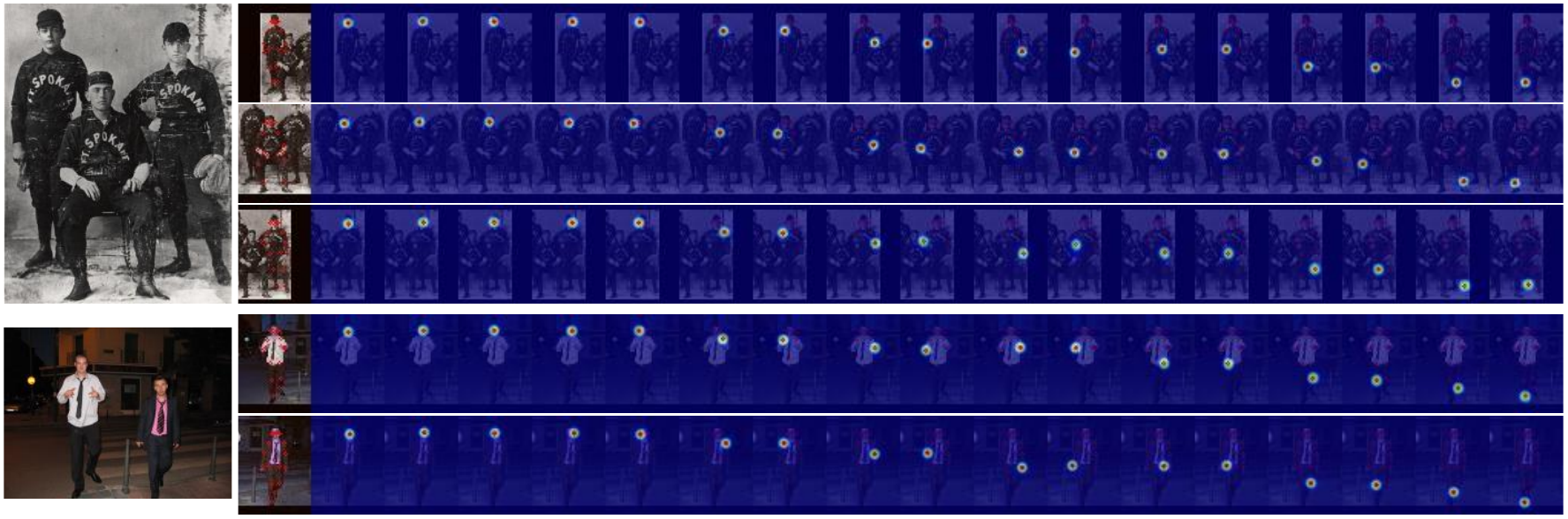}
  \caption{Illustration of the heatmaps of all persons in the image predicted by the network. Each heatmap of the same row denotes the location response of different keypoints of one person. It can be observed that in the presence of different human distractions, our method can still accurately estimate the location of key points in the human body.}
  \label{heatmap}
\end{figure*}

\subsection{Qualitative Results}
Our method incorporates both top-down and bottom-up pipelines, where the full-scene information of different visual fields can help capture human pose information, especially in complex scenes with multi-person interactions. Given an image, our network can accurately localize the location of keypoints for persons in the image. As shown in Fig.~\ref{qual_figs}, the pose estimation results of DPIT in different scenes are demonstrated. From the first row, we can observe that for complex pose scenes of a single person, our network can achieve accurate localization of keypoints. Benefitting from the fusion of visual information from different receptive fields and interactions of visual clues, the network still performs well while serious occlusion among different people. In addition, our method is not affected by the scale variation of the persons in the image. It is observable that accurate human pose estimation can still be reported for scenes with large-scale differences. As shown in Fig.~\ref{heatmap}, we further illustrate the heatmaps of keypoints, where the maximum response location represents the corresponding location of keypoints. As we can see, our network can predict precise heatmaps for different keypoints and different persons in the image. The heatmap manner effectively preserves the spatial location information.

\subsection{Ablation Studies}
In this section, to verify the contributions of different components and the influence of the structure parameter settings, we perform ablation experiments on the COCO dataset.

\subsubsection{Is it useful to employ the bottom-up branch to capture the full-scene information?}
\begin{figure*}
  \centerline{\includegraphics[width=0.8\textwidth]{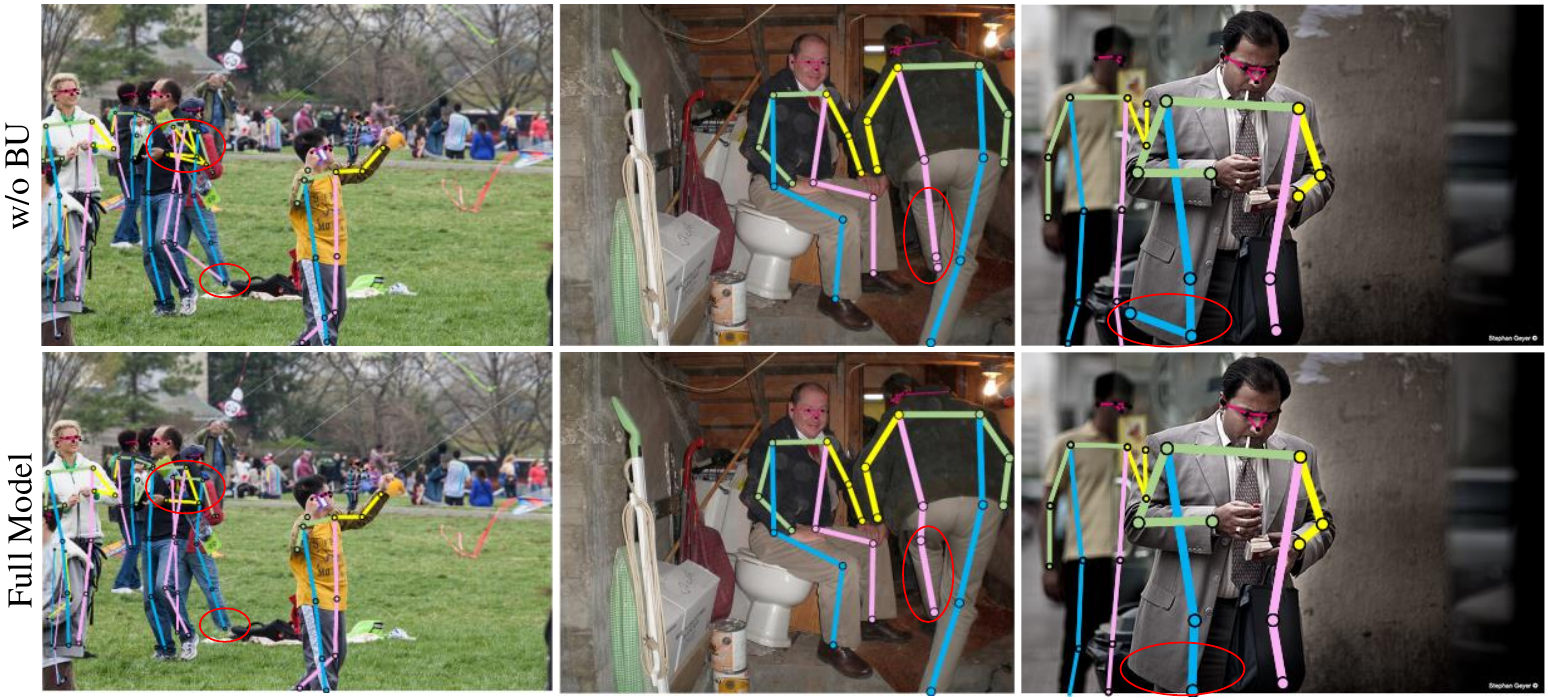}}
  \caption{Some qualitative results are illustrated to show the contributions of the bottom-up branch.}
  \label{ablation_fig}
\end{figure*}

The bottom-up branch extracts the full-scene information from the whole image with the encoder $\textbf{\textit{E}}_{}^{BU}$. To verify the usefulness of the bottom-up branch in locating human keypoints, we delete this branch in our DPIT, and obtain a network that does not utilize the feature of the entire image, called w/o $BU$. As shown in Table~\ref{ablation_table1}, we report quantitative results in the row with w/o $BU$. Without the help of the bottom-up branch, we observe degradation in the performance on different metrics. In addition, as shown in Fig.~\ref{ablation_fig}, the qualitative results indicate that the absence of full-scene information leads to inaccurate human pose estimation in the scenes with multi-person interaction and occlusion. As a result, both quantitative and qualitative results demonstrate the usefulness of the bottom-up branch.

\subsubsection{Why using transformer encoder?}
To evaluate the contribution of the transformer encoder to our DPIT, we perform another ablation experiment, i.e., removing the transformer encoder from the network. Alternatively, the featuremap of the bottom-up branch is integrated with the top-down branch by summation operation after some convolution layers, then the heatmap of the keypoints is predicted by the integrated features. It is worth mentioning that the simplified backbone no longer can capture enough information about the image in this case, which aggravates the network. In comparison, the transformer can capture long-term dependencies of the same visual field, while also integrating visual clues and posture clues from different perspectives with the help of self-attention mechanism. As shown in the row with w/o $Transformer$ of Table~\ref{ablation_table1}, the poorer performance of the network proves the effectiveness of the transformer.

\begin{minipage}[c]{0.5\textwidth}
\centering
\captionof{table}{Results on COCO validation set with the input size of $256 \times 192$. w/o $BU$: without bottom-up branch. w/o $Transformer$: without transformer.}
\label{ablation_table1}
\begin{tabular}{c|ccc}
  \toprule
  Model   & $AP$  & $AP^{50}$  & $AR$               \\
  \midrule
  \midrule
  w/o $BU$   & 76.6 & 92.5  & 79.4    \\
  w/o $Transformer$   & 76.5 & \textbf{93.6}  & 79.3    \\
  DPIT-B  & \textbf{76.9} & 93.5 & \textbf{79.6} \\
  \bottomrule
 \end{tabular}
\end{minipage}
\qquad
\begin{minipage}[c]{0.4\textwidth}
\centering
\captionof{table}{Ablation studies with different transformer encoder layers are performed on COCO validation dataset.}
\label{ablation_table2}
\begin{tabular}{c|c|ccc}
  \toprule
  Model   & Depth & $AP$ & $AP^{50}$  & $AR$ \\
  \midrule
  \midrule
  DPIT-B-D6  & 6  & 76.3 & 92.9 & 79.1    \\
  DPIT-B-D12 & 12 & \textbf{76.9} & \textbf{93.5} & \textbf{79.6}    \\
  DPIT-B-D16 & 16 & 76.5 & 92.6 & 79.3    \\
  \bottomrule
\end{tabular}
\end{minipage}
\\
\\
\subsubsection{What's the impact of depth of transformer encoder?}
For networks equipped with transformer, the depth of the encoder is a significant setting for the performance. As shown in Table~\ref{ablation_table2}, to explore the impact of the different number of encoder layers, we conduct quantitative experiments on the COCO validation set. Specifically, we only change the number of encoder layers of the transformer, keeping the other configurations fixed. Three different encoder layers are validated. It can be observed that different settings have different performances on the quantitative metrics. When the depth is shallow, the network performance improves as the depth increases. The metrics, however, also exhibit a decrease with too many encoder layers. Based on the
experimental validation, our final configuration of transformer encoder depth is 12 on COCO.

\section{Conclusion} \label{sec:Conclusion}
In this paper, we propose a novel Dual-Pipeline Integrated Transformer called DPIT for human pose estimation. To the best of our knowledge, this is one of the first works to integrate the bottom-up and top-down pipelines in one network with transformers. The proposed DPIT consists of two parts, a two-branch structure, and a feature interaction module. In our framework, the bottom-up branch and top-down branch capture full-scene information and posture visual clues with different receptive fields and perspectives, respectively. To achieve the effective integration of the two branches, the encoder of the transformer is applied to explore the long-term local-visual clues, global-visual clues, and their interactions. In addition, the defined keypoint embedding not only focuses on the different interest regions for a particular keypoint but also be allowed to concern the structural information between different keypoints. The reported quantitative and qualitative results on two public datasets demonstrate the effectiveness of our DPIT for human pose estimation.

\subsubsection{Acknowledgement.}
This research was supported by the National Key R\&D Program of China under Grant No. 2020AAA0103800.
\newpage

\begin{thebibliography}{10}
\providecommand{\url}[1]{\texttt{#1}}
\providecommand{\urlprefix}{URL }
\providecommand{\doi}[1]{https://doi.org/#1}

\bibitem{MPII20142d}
Andriluka, M., Pishchulin, L., Gehler, P., Schiele, B.: 2d human pose
  estimation: New benchmark and state of the art analysis. In: Proceedings of
  the IEEE Conference on computer Vision and Pattern Recognition. pp.
  3686--3693 (2014)

\bibitem{learning2020}
Cai, Y., Wang, Z., Luo, Z., Yin, B., Du, A., Wang, H., Zhang, X., Zhou, X.,
  Zhou, E., Sun, J.: Learning delicate local representations for multi-person
  pose estimation. In: European Conference on Computer Vision. pp. 455--472.
  Springer (2020)

\bibitem{RSN2020}
Cai, Y., Wang, Z., Luo, Z., Yin, B., Du, A., Wang, H., Zhang, X., Zhou, X.,
  Zhou, E., Sun, J.: Learning delicate local representations for multi-person
  pose estimation. In: European Conference on Computer Vision. pp. 455--472.
  Springer (2020)

\bibitem{0penpose2017}
Cao, Z., Simon, T., Wei, S.E., Sheikh, Y.: Realtime multi-person 2d pose
  estimation using part affinity fields. In: Proceedings of the IEEE conference
  on computer vision and pattern recognition. pp. 7291--7299 (2017)

\bibitem{DETR2020}
Carion, N., Massa, F., Synnaeve, G., Usunier, N., Kirillov, A., Zagoruyko, S.:
  End-to-end object detection with transformers. In: European conference on
  computer vision. pp. 213--229. Springer (2020)

\bibitem{CPN2018}
Chen, Y., Wang, Z., Peng, Y., Zhang, Z., Yu, G., Sun, J.: Cascaded pyramid
  network for multi-person pose estimation. In: Proceedings of the IEEE
  conference on computer vision and pattern recognition. pp. 7103--7112 (2018)

\bibitem{Higherhrnet2020}
Cheng, B., Xiao, B., Wang, J., Shi, H., Huang, T.S., Zhang, L.: Higherhrnet:
  Scale-aware representation learning for bottom-up human pose estimation. In:
  Proceedings of the IEEE/CVF conference on computer vision and pattern
  recognition. pp. 5386--5395 (2020)

\bibitem{VIT2020}
Dosovitskiy, A., Beyer, L., Kolesnikov, A., Weissenborn, D., Zhai, X.,
  Unterthiner, T., Dehghani, M., Minderer, M., Heigold, G., Gelly, S., et~al.:
  An image is worth 16x16 words: Transformers for image recognition at scale.
  arXiv preprint arXiv:2010.11929  (2020)

\bibitem{RMPE2017}
Fang, H.S., Xie, S., Tai, Y.W., Lu, C.: Rmpe: Regional multi-person pose
  estimation. In: Proceedings of the IEEE international conference on computer
  vision. pp. 2334--2343 (2017)

\bibitem{DPM2010}
Felzenszwalb, Pedro, F., Girshick, Ross, B., McAllester, David, Ramanan, Deva:
  Object detection with discriminatively trained part-based models. IEEE
  Transactions on Pattern Analysis \& Machine Intelligence  \textbf{32}(9),
  1627--1645 (2010)

\bibitem{DEKR2021}
Geng, Z., Sun, K., Xiao, B., Zhang, Z., Wang, J.: Bottom-up human pose
  estimation via disentangled keypoint regression. In: Proceedings of the
  IEEE/CVF Conference on Computer Vision and Pattern Recognition. pp.
  14676--14686 (2021)

\bibitem{latent2011}
Ionescu, C., Li, F., Sminchisescu, C.: Latent structured models for human pose
  estimation. In: 2011 International Conference on Computer Vision. pp.
  2220--2227. IEEE (2011)

\bibitem{2014Adam}
Kingma, D., Ba, J.: Adam: A method for stochastic optimization. Computer
  Science  (2014)

\bibitem{MSPN2019}
Li, W., Wang, Z., Yin, B., Peng, Q., Du, Y., Xiao, T., Yu, G., Lu, H., Wei, Y.,
  Sun, J.: Rethinking on multi-stage networks for human pose estimation. arXiv
  preprint arXiv:1901.00148  (2019)

\bibitem{Tokenpose2021}
Li, Y., Zhang, S., Wang, Z., Yang, S., Yang, W., Xia, S.T., Zhou, E.:
  Tokenpose: Learning keypoint tokens for human pose estimation. In:
  Proceedings of the IEEE/CVF International Conference on Computer Vision. pp.
  11313--11322 (2021)

\bibitem{COCO2014microsoft}
Lin, T.Y., Maire, M., Belongie, S., Hays, J., Perona, P., Ramanan, D.,
  Doll{\'a}r, P., Zitnick, C.L.: Microsoft coco: Common objects in context. In:
  European conference on computer vision. pp. 740--755. Springer (2014)

\bibitem{liu2018t}
Liu, K., Liu, W., Gan, C., Tan, M., Ma, H.: T-c3d: Temporal convolutional 3d
  network for real-time action recognition. In: Proceedings of the AAAI
  conference on artificial intelligence. vol.~32 (2018)

\bibitem{liu2020real}
Liu, K., Liu, W., Ma, H., Tan, M., Gan, C.: A real-time action representation
  with temporal encoding and deep compression. IEEE Transactions on Circuits
  and Systems for Video Technology  \textbf{31}(2),  647--660 (2020)

\bibitem{swin-transformer2021}
Liu, Z., Lin, Y., Cao, Y., Hu, H., Wei, Y., Zhang, Z., Lin, S., Guo, B.: Swin
  transformer: Hierarchical vision transformer using shifted windows. In:
  Proceedings of the IEEE/CVF International Conference on Computer Vision. pp.
  10012--10022 (2021)

\bibitem{Tfpose2021}
Mao, W., Ge, Y., Shen, C., Tian, Z., Wang, X., Wang, Z.: Tfpose: Direct human
  pose estimation with transformers. arXiv preprint arXiv:2103.15320  (2021)

\bibitem{associative2017}
Newell, A., Huang, Z., Deng, J.: Associative embedding: End-to-end learning for
  joint detection and grouping. Advances in neural information processing
  systems  \textbf{30} (2017)

\bibitem{Hourglass2016}
Newell, A., Yang, K., Deng, J.: Stacked hourglass networks for human pose
  estimation. In: European conference on computer vision. pp. 483--499.
  Springer (2016)

\bibitem{G-RMI2017}
Papandreou, G., Zhu, T., Kanazawa, N., Toshev, A., Tompson, J., Bregler, C.,
  Murphy, K.: Towards accurate multi-person pose estimation in the wild. In:
  2017 IEEE Conference on Computer Vision and Pattern Recognition (CVPR) (2017)

\bibitem{strong2013}
Pishchulin, L., Andriluka, M., Gehler, P., Schiele, B.: Strong appearance and
  expressive spatial models for human pose estimation. In: Proceedings of the
  IEEE international conference on Computer Vision. pp. 3487--3494 (2013)

\bibitem{deepcut2016}
Pishchulin, L., Insafutdinov, E., Tang, S., Andres, B., Andriluka, M., Gehler,
  P.V., Schiele, B.: Deepcut: Joint subset partition and labeling for multi
  person pose estimation. In: Proceedings of the IEEE conference on computer
  vision and pattern recognition. pp. 4929--4937 (2016)

\bibitem{randomized2008}
Rogez, G., Rihan, J., Ramalingam, S., Orrite, C., Torr, P.H.: Randomized trees
  for human pose detection. In: 2008 IEEE Conference on Computer Vision and
  Pattern Recognition. pp.~1--8. IEEE (2008)

\bibitem{ImageNet2015}
Russakovsky, O., Deng, J., Su, H., Krause, J., Satheesh, S., Ma, S., Huang, Z.,
  Karpathy, A., Khosla, A., Bernstein, M., et~al.: Imagenet large scale visual
  recognition challenge. International journal of computer vision
  \textbf{115}(3),  211--252 (2015)

\bibitem{POET2021}
Stoffl, L., Vidal, M., Mathis, A.: End-to-end trainable multi-instance pose
  estimation with transformers. arXiv preprint arXiv:2103.12115  (2021)

\bibitem{segmenter2021}
Strudel, R., Garcia, R., Laptev, I., Schmid, C.: Segmenter: Transformer for
  semantic segmentation. In: Proceedings of the IEEE/CVF International
  Conference on Computer Vision. pp. 7262--7272 (2021)

\bibitem{HRNet2019}
Sun, K., Xiao, B., Liu, D., Wang, J.: Deep high-resolution representation
  learning for human pose estimation. In: Proceedings of the IEEE/CVF
  Conference on Computer Vision and Pattern Recognition. pp. 5693--5703 (2019)

\bibitem{DEiT2021}
Touvron, H., Cord, M., Douze, M., Massa, F., Sablayrolles, A., J{\'e}gou, H.:
  Training data-efficient image transformers \& distillation through attention.
  In: International Conference on Machine Learning. pp. 10347--10357. PMLR
  (2021)

\bibitem{sparse2008}
Urtasun, R., Darrell, T.: Sparse probabilistic regression for
  activity-independent human pose inference. In: 2008 IEEE Conference on
  Computer Vision and Pattern Recognition. pp.~1--8. IEEE (2008)

\bibitem{transformer2017}
Vaswani, A., Shazeer, N., Parmar, N., Uszkoreit, J., Jones, L., Gomez, A.N.,
  Kaiser, {\L}., Polosukhin, I.: Attention is all you need. Advances in neural
  information processing systems  \textbf{30} (2017)

\bibitem{graph2020}
Wang, J., Long, X., Gao, Y., Ding, E., Wen, S.: Graph-pcnn: Two stage human
  pose estimation with graph pose refinement. In: European Conference on
  Computer Vision. pp. 492--508. Springer (2020)

\bibitem{sceneformer2021}
Wang, X., Yeshwanth, C., Nie{\ss}ner, M.: Sceneformer: Indoor scene generation
  with transformers. In: 2021 International Conference on 3D Vision (3DV). pp.
  106--115. IEEE (2021)

\bibitem{CPM2016}
Wei, S.E., Ramakrishna, V., Kanade, T., Sheikh, Y.: Convolutional pose
  machines. In: Proceedings of the IEEE conference on Computer Vision and
  Pattern Recognition. pp. 4724--4732 (2016)

\bibitem{simple_baseline2018}
Xiao, B., Wu, H., Wei, Y.: Simple baselines for human pose estimation and
  tracking. In: Proceedings of the European conference on computer vision
  (ECCV). pp. 466--481 (2018)

\bibitem{segformer2021}
Xie, E., Wang, W., Yu, Z., Anandkumar, A., Alvarez, J.M., Luo, P.: Segformer:
  Simple and efficient design for semantic segmentation with transformers.
  Advances in Neural Information Processing Systems  \textbf{34} (2021)

\bibitem{Swinpose2022}
Xiong, Z., Wang, C., Li, Y., Luo, Y., Cao, Y.: Swin-pose: Swin transformer
  based human pose estimation. arXiv preprint arXiv:2201.07384  (2022)

\bibitem{Transpose2021}
Yang, S., Quan, Z., Nie, M., Yang, W.: Transpose: Keypoint localization via
  transformer. In: Proceedings of the IEEE/CVF International Conference on
  Computer Vision. pp. 11802--11812 (2021)

\bibitem{Simple2021}
Zhang, J., Zhu, Z., Lu, J., Huang, J., Huang, G., Zhou, J.: Simple:
  Single-network with mimicking and point learning for bottom-up human pose
  estimation. arXiv preprint arXiv:2104.02486  (2021)

\bibitem{deformable2020}
Zhu, X., Su, W., Lu, L., Li, B., Wang, X., Dai, J.: Deformable detr: Deformable
  transformers for end-to-end object detection. arXiv preprint arXiv:2010.04159
   (2020)

\end{thebibliography}
%

\end{document}